# Optimal Selection of Measurement Configurations for Stiffness Model Calibration of Anthropomorphic Manipulators

Alexandr Klimchik[1, 2, a], Yier Wu[1, 2, b],
Anatol Pashkevich[1, 2, c], Stéphane Caro[1, 2, d], Benoît Furet[1, 2, e]

[1] Ecole des Mines de Nantes, 4 rue Alfred-Kastler, 44307 Nantes, France
[2] Institut de Recherche en Communications et Cybernétique de Nantes, 1 rue de la Noë, France

[a] alexandr.klimchik@mines-nantes.fr (corresponding author), [b] yier.wu@mines-nantes.fr ,
[c] anatol.pashkevich@mines-nantes.fr, [d] stephane.caro@irccyn.ec-nantes.fr,
[e] benoit.furet@irccyn.ec-nantes.fr

**Abstract.** The paper focuses on the calibration of elastostatic parameters of spatial anthropomorphic robots. It proposes a new strategy for optimal selection of the measurement configurations that essentially increases the efficiency of robot calibration. This strategy is based on the concept of the robot test-pose and ensures the best compliance error compensation for the test configuration. The advantages of the proposed approach and its suitability for practical applications are illustrated by numerical examples, which deal with calibration of elastostatic parameters of a 3 degrees of freedom anthropomorphic manipulator with rigid links and compliant actuated joints.

**Introduction**

In the usual engineering practice, the accuracy of an anthropomorphic manipulator depends on a number of factors. Following [1-2], the main sources of robot positioning errors can be divided into two principal groups: geometrical (link lengths, assembling errors, errors in the joint zero values et al.) and non-geometrical ones (compliant errors, measurement errors, environment factors, control errors, friction, backlash, wear et al.). For the industrial manipulators, the most essential of them are related to the manufacturing tolerances leading to the geometrical parameters deviation with respect to their nominal values (the geometrical errors) as well as to the end-effector deflections caused by the applied forces and torques (the compliance errors). It is worth mentioning that these sources of errors may be either independent or correlated, but, in practice, they are usually treated sequentially, assuming that they are statistically independent.

Usually, for the industrial applications where the external forces/torques applied to the end-effector are relatively small, the prime source of the manipulator inaccuracy is the *geometrical errors*. As reported by several authors [3], they are responsible for about 90% of the total position error. These errors are associated with the differences between the nominal and actual values of the link/joint parameters. Typical examples of them are the differences between the nominal and the actual length of links, the differences between zero values of actuator coordinates in the real robot and the mathematical model embedded in the controller (joint offsets) [4]. They can be also induced by the non-perfect assembling of different links and lead to shifting and/or rotation of the frames associated with different elements, which are normally assumed to be matched and aligned. It is clear that the geometrical errors do not depend on the manipulator configuration, while their effect on the position accuracy depends on the last one. At present, there exists various sophisticated calibration techniques that are able to identify the differences between the actual and the nominal geometrical parameters [5-9]. Consequently, this type of errors can be efficiently compensated either by adjusting the controller input (i.e. the target point coordinates) or by straightforward modification of the geometrical model parameters used in the robot controller.

In some other cases, the geometrical errors may be dominated by *non-geometrical* ones that may be caused by influences of a number of factors [10-11]. However, in the regular service conditions, the *compliance errors* are the most significant source of inaccuracy. Their influence is particularly important for heavy robots and for manipulators with low stiffness. For example, the cutting forces/torques from the technological process may induce significant deformations, which are not negligible in the precise machining. In this case, the influence of the compliance errors on the robot position accuracy can be even higher than the geometrical ones.

Generally, the compliance errors depend on two main factors: (i) the stiffness of the manipulator and (ii) the loading applied to it. Similar to the geometrical ones, the compliance errors highly depend on the manipulator configuration and essentially differ throughout the workspace [12]. So, in order to obtain correct prediction of the robot end-effector position, the maximum compliance errors compensation should be achieved [13]. One way to solve this problem is to improve the accuracy of the stiffness model by means of elastostatic calibration. This procedure allows to identify the stiffness parameters from the redundant information on the state of the robot end-effector position provided by the measurements, where the impacts of associated measurement noise on the calibration results have to be minimized.

However, currently most of the efforts have been made for kinematic calibration, only few works directly address the issue of elastostatic calibration and its influences on the robot accuracy [14]. Besides, using various manipulator configurations for different measurements seems to be attractive and perfectly corresponds to some basic ideas of the classical design of experiments theory [15] that intends using the factors that are differed from each other as much as possible. In spite of potential advantages of this approach and potential benefits to improve the identification accuracy significantly, only few works addressed to the issue of the best measurement pose selection [16-19]. Hence, the problem of selection of the optimal measurement poses for elastostatic parameters calibration requires additional investigation. This problem can be treated as finding the strategy of determining a set of optimal measurement poses within the reachable joint space that minimize the effects of measurement noise on the estimation of the robot parameters. It should be mentioned that the end-effector location as well as its deflection under the loading are described by a non-linear set of functions. However, the classical results of the identification theory are mostly obtained for very specific models (such as linear regression), Therefore, they can not be applied directly and an additional enhancement is required.

One of the key issues in the experiment design theory is comparison of the experimental plans. In the literature, in order to define the optimal experimental plan, numerous quantitative performance measures that reduce multi-objective optimization problem to a scalar factor have been proposed. Consequently, different factors that evaluate robot calibration performance have been defined as the objectives of optimization, associated with a set of measurement poses [20-24]. However, all the existing factors have their limitations that affect the calibration accuracy in different manners. As a result, they do not entirely correspond to the industrial requirements. This motivates a research direction of this work.

In this paper, the problem of optimal design of the elastostatic calibration experiments is studied for the case of 3-link spatial anthropomorphic manipulator, which obviously does not cover all architectures used in practice. Nevertheless, it allows us to derive very useful analytical expressions and to obtain some simple practical rules defining optimal configurations with respect to the calibration accuracy. In contrast to other works, it is proposed a new criterion that evaluates the quality of compliance errors compensation based on the concept of manipulator test-pose. The proposed criterion has a clear physical meaning and directly related to the robot accuracy, and allows us essentially improving the efficiency of compliance errors compensation via proper selection of measurement poses.

**Problem statement**

The elastostatic properties of a serial robotic manipulator [12] are usually defined by Cartesian stiffness matrix $\mathbf{K}_C$, which is computed as

$$\mathbf{K}_C = \mathbf{J}^{-T} \mathbf{K}_\theta \mathbf{J}^{-1} \qquad (1)$$

where $\mathbf{J}$ is the Jacobian matrix with respect to the joint angles $\mathbf{q}$, and $\mathbf{K}_\theta$ is a diagonal matrix that aggregates stiffness of the joints. In order to describe the linear relation between the end-effector displacement and the external force, the stiffness model of this manipulator can be rewritten as follows

$$\Delta \mathbf{p} = \mathbf{J}\, \mathbf{k}_\theta \mathbf{J}^T \mathbf{F} \qquad (2)$$

where $\Delta \mathbf{p}$ is the robot end-effector displacement caused by the external loading, $\mathbf{k}_\theta$ is the joints compliance matrix; $\mathbf{F}$ is the external force/torque.

It is assumed that the geometric parameters are well calibrated. So, for the unloaded mode ($\mathbf{F} = 0$), the vector $\mathbf{q}$ is equal to the nominal value of the joint angles $\mathbf{q}_0$. However, for the case when the loading is not equal to zero $\mathbf{F} \neq 0$, the joint angles include deflections, i.e. $\mathbf{q} = \mathbf{q}_0 + \delta \mathbf{q}$, where $\delta \mathbf{q}$ is the vector of joint displacements due to the external loading $\mathbf{F}$. Thus, the elastostatic model (2) includes parameters of $\mathbf{k}_\theta$ that must be identified by means of calibration.

It is assumed that each calibration experiment produces three vectors $\{\Delta \mathbf{p}_i, \mathbf{q}_i, \mathbf{F}_i\}$, which define the displacements of the robot end-effector, the corresponding joint angles and the external forces respectively, where $i$ is the experiment number. So, the calibration procedure may be treated as the best fitting of the experimental data $\{\Delta \mathbf{p}_i, \mathbf{q}_i, \mathbf{F}_i\}$ by using the stiffness model (2) that can be solved using the standard least-square technique.

In practice, the calibration includes measurements of the end-effector Cartesian coordinates with some errors, which are assumed to be i.i.d (independent identically distributed) random values with zero expectation and standard deviation $\sigma$. Because of these errors, the desired values of $\mathbf{k}_\theta$ are always identified approximately. So, the problem of interest is to evaluate the identification accuracy for the desired parameters and to propose a technique for selecting the set of joint variables $\mathbf{q}_i$ and external forces $\mathbf{F}_i$ that leads to the accuracy improvement.

Usually, the performance measures that evaluate the quality of the calibration plans are based on the analyses of the covariance matrix of the identified parameters, all elements of which should be as small as possible. However, in robots the stiffness parameters $(k_1, k_2, ...)$ have different influences on the end-effector displacements; moreover, their influence varies throughout the workspace. To overcome this difficulty, in this work it is assumed that the "calibration quality" is evaluated for the so-called test configuration $\{\mathbf{q}^0, \mathbf{F}^0\}$, which is given by a user and for which it is required to have the best positioning accuracy under external loading.

To solve this general problem, two sub-problems should be considered: (i) to propose a optimality criterion that is adapted to the elstostatic parameters calibration of the anthropomorphic manipulator; (ii) to find optimal configurations of the manipulator for elastostatic parameters calibration that provide the best compensation of errors.

**Influence of measurement errors**

For computational convenience, the linear relation (2) where the desired parameters are arranged in the diagonal matrix $\mathbf{k}_\theta = diag(k_1, k_2, ...)$ should be rewritten in the following form

$$\Delta \mathbf{p}_i = \mathbf{A}_i \mathbf{k} \qquad (3)$$

where the vector $\mathbf{k}$ collects the joint compliances that are extracted from matrix $\mathbf{k}_\theta$; the matrix $\mathbf{A}_i$ is defined by the columns of Jacobian $\mathbf{J}$ and the external force $\mathbf{F}$ and is expressed as

$$\mathbf{A}_i = \left[ \mathbf{J}_{1i} \mathbf{J}_{1i}^T \mathbf{F}_i \;\vdots\; \mathbf{J}_{2i} \mathbf{J}_{2i}^T \mathbf{F}_i \;\vdots\; \cdots \;\vdots\; \mathbf{J}_{ni} \mathbf{J}_{ni}^T \mathbf{F}_i \right] \quad (i = \overline{1, m}) \qquad (4)$$

where $\mathbf{J}_{ni}$ is the $n^{th}$ column vector of the Jacobian matrix for the $i^{th}$ experiment, $m$ is the number of experiments. Using the identification theory, the joint compliances can be obtained from Eq. (3) using least square method, which minimizes the residuals for all experimental data. The corresponding optimization problem can be formulated as

$$\sum_{i=1}^{m} (\mathbf{A}_i \mathbf{k} - \Delta \mathbf{p}_i)^T (\mathbf{A}_i \mathbf{k} - \Delta \mathbf{p}_i) \to \min_{\mathbf{q}_i, \mathbf{F}_i} \quad (5)$$

The solution of this optimization problem provides the estimation of desired parameters, which can be computed as

$$\mathbf{k}_0 = \left( \sum_{i=1}^{m} \mathbf{A}_i^T \mathbf{A}_i \right)^{-1} \cdot \left( \sum_{i=1}^{m} \mathbf{A}_i^T \Delta \mathbf{p}_i \right) \quad (6)$$

Considering that in the calibration experiments the measurement errors cannot be avoided, Eq. (3) should be rewritten in the following form

$$\Delta \mathbf{p}_i = \mathbf{A}_i \mathbf{k} + \varepsilon_i \quad (7)$$

where $\varepsilon_i$ is the measurement errors in the $i^{th}$ experiment with the expectation $\mathrm{E}(\varepsilon_i) = 0$ and the variance $\mathrm{E}(\varepsilon_i^T \varepsilon_i) = \sigma^2$. It is evident that the measurement errors have affects on the identification accuracy of the unknown parameters $\mathbf{k}$. So, the estimation of desired parameters $\mathbf{k}$ takes the form

$$\mathbf{k} = \left( \sum_{i=1}^{m} \mathbf{A}_i^T \mathbf{A}_i \right)^{-1} \left( \sum_{i=1}^{m} \mathbf{A}_i^T (\Delta \mathbf{p}_i - \varepsilon_i) \right) \quad (8)$$

As follows from (8), the latter expression produces unbiased estimates $\mathrm{E}(\mathbf{k}) = \mathbf{k}_0$. It can be also proved that the covariance matrix of compliance parameters (8) that defines the identification accuracy can be expressed as

$$\mathrm{cov}(\mathbf{k}) = \left( \sum_{i=1}^{m} \mathbf{A}_i^T \mathbf{A}_i \right)^{-1} \mathrm{E}\left( \sum_{i=1}^{m} \mathbf{A}_i^T \varepsilon_i^T \varepsilon_i \mathbf{A}_i \right) \left( \sum_{i=1}^{m} \mathbf{A}_i^T \mathbf{A}_i \right)^{-1} \quad (9)$$

Then, taking into account that $\mathrm{E}\left( \sum_{i=1}^{m} \varepsilon_i^T \varepsilon_i \right) = \sigma^2 \mathbf{I}$, where $\mathbf{I}$ is $n \times n$ identity matrix, Eq. (9) can be simplified to

$$\mathrm{cov}(\mathbf{k}) = \sigma^2 \left( \sum_{i=1}^{m} \mathbf{A}_i^T \mathbf{A}_i \right)^{-1} \quad (10)$$

where $\sigma$ is the standard deviation of the measurement errors. So, for the considered problem, the impact of the measurement errors is defined by the matrix sum $\sum_{i=1}^{m} \mathbf{A}_i^T \mathbf{A}_i$ that is also called the information matrix.

Obviously, in order to have the smallest dispersion of the identification errors, it is required to have the covariance matrix elements as small as possible. It is a multiobjective optimization problem, but minimization of one element can possibly increase others. So, in order to reduce this problem to a nonobjective one, numerous scalar criteria have been proposed. It should be mentioned that all these criteria provide rather different optimal solutions. So, it is quite important to select a proper optimization criteria that ensures the best position accuracy of the manipulator under the loading. For this reason, in the next section a new test-pose based approach that ensures the best end-effector accuracy under external loading is proposed.

**Test-pose based approach for calibration of elastostatic parameters**

In order to give more clear physical meaning related to the robot accuracy, a new optimality criterion is proposed to evaluate the mean squared error of the joint compliances (end-effector deflections) for a given test pose. It evaluates the ability to compensate the compliance errors for given test pose. Similar approach for geometrical calibration has been used in [25].

Assuming that the measurement errors have affects on the identification accuracy, Eq. (3) can be expressed in a different manner:

$$\Delta \mathbf{p} + \delta \mathbf{p} = \mathbf{A}^0 (\mathbf{k} + \delta \mathbf{k}) \qquad (11)$$

where $\delta \mathbf{p}$ stands for the deflection error, and $\delta \mathbf{k}$ describes the compliance parameter error; the matrix $\mathbf{A}^0$ is defined by the given test pose using (4). Taking into account that $\Delta \mathbf{p} = \mathbf{A}^0 \mathbf{k}$, Eq. (11) is equivalent to

$$\delta \mathbf{p} = \mathbf{A}^0 \cdot \delta \mathbf{k} \qquad (12)$$

So, the mean squared error of the joint compliances under the external loading, can be expressed as

$$O_t = \mathrm{E}(\delta \mathbf{p}^T \delta \mathbf{p}) = \mathrm{E}(\delta \mathbf{k}^T \mathbf{A}^{0T} \mathbf{A}^0 \delta \mathbf{k}) \qquad (13)$$

In order to simplify equation (13), it is possible to replace the term $\delta \mathbf{p}^T \delta \mathbf{p}$, by $\mathrm{trace}(\delta \mathbf{p} \delta \mathbf{p}^T)$, then

$$O_t = \mathrm{trace}\left(\mathrm{E}\left(\mathbf{A}^0 \delta \mathbf{k} \delta \mathbf{k}^T \mathbf{A}^{0T}\right)\right) \qquad (14)$$

Since $\mathrm{E}(\delta \mathbf{k} \delta \mathbf{k}^T)$ is the covariance matrix of desired parameters $\mathbf{k}$, the proposed performance measure (14) can be presented as

$$O_t = \sigma^2 \mathrm{trace}\left(\mathbf{A}^0 \left(\sum_{i=1}^m \mathbf{A}_i^T \mathbf{A}_i\right)^{-1} \mathbf{A}^{0T}\right) \qquad (15)$$

**Remark 1** The proposed criterion can be treated as the weighted trace of the covariance matrix of the desired parameters, where the weighting coefficients are derived using the test pose.

**Remark 2** If the test pose and measurement poses are the same, which means that $\mathbf{A}_i = \mathbf{A}^0$, the s.t.d. of the compensation errors can be expressed as

$$\left. O_t \right|_{\mathbf{A}_i = \mathbf{A}_t, i=\overline{1,m}} = \frac{n \sigma^2}{m} \qquad (16)$$

where $n$ is the number of identifiable parameters and $m$ is the number of measurements (it is obviously an upper bound that should be reduced by a proper selection of the measurement poses).

Hence, the proposed optimization criterion ensures low values of the covariance matrix elements and allows to combine multiple objectives with different units in a single scalar factor. An application of this criterion for a proper selection of the measurement poses for the spatial anthropomorphic robot is illustrated in the next section.

**Elastostatic parameters calibration for an anthropomorphic manipulator**

Let us consider the problem of optimal configuration selection for calibration of the elastostatic parameters of a 3-link spatial anthropomorphic manipulator with rigid links and compliant actuators (Fig. 1). The geometrical model of the considered robot can be defined as

$$x = (l_2 cos(q_2) + l_3 cos(q_2 + q_3)) cos(q_1)$$
$$y = (l_2 cos(q_2) + l_3 cos(q_2 + q_3)) sin(q_1) \quad (17)$$
$$z = l_1 + l_2 sin(q_2) + l_3 sin(q_2 + q_3)$$

where $l_1, l_2, l_3$ denote the link lengths and the joint angles $q_1, q_2, q_3$ characterize manipulator configuration $\mathbf{q}$. It is assumed that this manipulator should execute a prescribed task in the configuration $\mathbf{q}^0 = (q_1^0, q_2^0, q_3^0)$ under payload $\mathbf{F}^0 = (F_x^0 \quad F_y^0 \quad F_z^0)^T$ with a high precision (here superscript "0" denotes the test configuration). Besides, it is also assumed that the geometrical model is accurate, geometrical parameters are well calibrated (errors in geometrical parameters can be neglected). Hence, in order to ensure high accuracy for the error compensation, it is required to identify the joint compliances $k_1, k_2, k_3$. To estimate the quality of the robot calibration process (defined by the set of configurations chosen for the measurement), let us use the test-pose based criterion (15) that improves the efficiency of the error compensation.

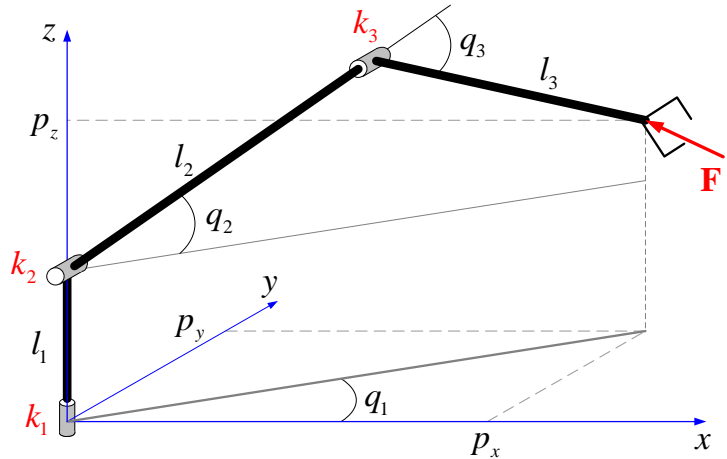

Fig. 1    3-link spatial anthropomorphic manipulator

For the considered 3-link robot, Jacobian matrix for the test pose $\mathbf{q}^0 = (q_1^0, q_2^0, q_3^0)$ can be written in the following form

$$\mathbf{J}^0 = \begin{bmatrix} -l_C^0 sin(q_1^0) & -l_S^0 cos(q_1^0) & -l_3 sin(q_2^0 + q_3^0) cos(q_1^0) \\ l_C^0 cos(q_1^0) & -l_S^0 sin(q_1^0) & -l_3 sin(q_2^0 + q_3^0) sin(q_1^0) \\ 0 & l_C^0 & l_3 cos(q_2^0 + q_3^0) \end{bmatrix} \quad (18)$$

where

$$l_C^0 = l_2 cos(q_2^0) + l_3 cos(q_2^0 + q_3^0); \qquad l_S^0 = l_2 sin(q_2^0) + l_3 sin(q_2^0 + q_3^0) \quad (19)$$

Using this expression, the matrix $\mathbf{A}^0$ for the test configuration can be expressed in the matrix (see Eq. (4)) form via vector columns as

$$\mathbf{A}^0 = \begin{bmatrix} \mathbf{A}_1^0 & \mathbf{A}_2^0 & \mathbf{A}_3^0 \end{bmatrix} \quad (20)$$

where the vectors $\mathbf{A}_1^0, \mathbf{A}_2^0, \mathbf{A}_3^0$ are defined as

$$\begin{aligned}
\mathbf{A}_1^0 &= \left(l_C^0\right)^2 \left(F_x^0 \sin(q_1^0) - F_y^0 \cos(q_1^0)\right) \begin{bmatrix} \sin(q_1^0) & -\cos(q_1^0) & 0 \end{bmatrix}^T \\
\mathbf{A}_2^0 &= \left(F_x^0 l_S^0 \cos(q_1^0) + F_y^0 l_S^0 \cdot \sin(q_1^0) - F_z^0 l_C^0\right) \begin{bmatrix} l_S^0 \cos(q_1^0) & l_S^0 \sin(q_1^0) & -l_C^0 \end{bmatrix}^T \\
\mathbf{A}_3^0 &= \left(F_x^0 l_3 \sin(q_2^0 + q_3^0)\cos(q_1^0) + F_y^0 l_3 \sin(q_2^0 + q_3^0)\sin(q_1^0) - F_z^0 l_3 \cos(q_2^0 + q_3^0)\right) \cdot \\
&\quad \cdot \begin{bmatrix} l_3 \sin(q_2^0 + q_3^0)\cos(q_1^0) & l_3 \sin(q_2^0 + q_3^0)\sin(q_1^0) & -l_3 \cos(q_2^0 + q_3^0) \end{bmatrix}^T
\end{aligned} \qquad (21)$$

In order to reduce the number of optimization parameters in the posture (some of them are obviously redundant), so it is reasonable to consider calibration configurations with $q_{1i}$ equal to zero (here, subscript "$i$" defines the experiment number). So, the Jacobian for $i^{th}$ experiment can be expressed as

$$\mathbf{J}_i = \begin{bmatrix} 0 & -l_2 \sin(q_{2i}) - l_3 \sin(q_{2i} + q_{3i}) & -l_3 \sin(q_{2i} + q_{3i}) \\ l_2 \cos(q_{2i}) + l_3 \cos(q_{2i} + q_{3i}) & 0 & 0 \\ 0 & l_2 \cos(q_{2i}) + l_3 \cos(q_{2i} + q_{3i}) & l_3 \cos(q_{2i} + q_{3i}) \end{bmatrix} \qquad (22)$$

Another redundant variable is $F_{xi}$, it can be taken into account by $F_{zi}$ and angle $q_{2i}$. Therefore, without loss of generality, force $\mathbf{F}_i$ can take the form

$$\mathbf{F}_i = \begin{bmatrix} 0 & F_0 \cos(\alpha_i) & F_0 \sin(\alpha_i) \end{bmatrix}^T \qquad (23)$$

where $F_0$ defines the force magnitude, which is suppose to be the same for all experiments and the angle $\alpha_i$ defines the force orientation in the yz plane. Under such assumptions, the term $F_{yi}$ causes deformations in the first joint and the term $F_{zi}$ causes deformations in the second and the third joints.

Using (22) and (23) the matrix $\mathbf{A}_i$, defined in Eq. (4), for the i-th experiment can be expressed as

$$\mathbf{A}_i = F_0 \begin{bmatrix} 0 & -l_{Si} l_{Ci} \sin(\alpha_i) & -l_3^2 \cos(q_{2i} + q_{3i}) \sin(q_{2i} + q_{3i}) \sin(\alpha_i) \\ l_{Ci}^2 \cos(\alpha_i) & 0 & 0 \\ 0 & l_{Ci}^2 \sin(\alpha_i) & l_3^2 \cos^2(q_{2i} + q_{3i}) \sin(\alpha_i) \end{bmatrix} \qquad (24)$$

where $l_{Ci}$ and $l_{Si}$ can be computed similar to (19). So, the information matrix can be computed as

$$\sum_{i=1}^m \mathbf{A}^T \mathbf{A} = F_0^2 \begin{bmatrix} a_{11} & 0 & 0 \\ 0 & a_{22} & a_{23} \\ 0 & a_{23} & a_{33} \end{bmatrix} \qquad (25)$$

where $m$ is the number of experiments and $a_{11}, a_{22}, a_{33}, a_{23}$ are expressed as

$$\begin{aligned}
a_{11} &= \sum_{i=1}^m l_{Ci}^4 \cos^2(\alpha_i); & a_{22} &= \sum_{i=1}^m l_{Ci}^2 \left(l_2^2 + l_3^2 + 2l_2 l_3 \cos(q_{3i})\right) \sin^2(\alpha_i) \\
a_{33} &= \sum_{i=1}^m l_3^4 \cos^2(q_{2i} + q_{3i}) \sin^2(\alpha_i); & a_{23} &= \sum_{i=1}^m l_3^2 l_{Ci} \cos(q_{2i} + q_{3i})\left(l_3 + l_2 \cos(q_{3i})\right) \sin^2(\alpha_i)
\end{aligned} \qquad (26)$$

Hence, for the considered robot, the covariance matrix $\text{cov}(\mathbf{k})$ can be expressed as

$$\text{cov}(\mathbf{k}) = \frac{\sigma^2}{F_0^2} \cdot \begin{bmatrix} 1/a_{11} & 0 & 0 \\ 0 & a_{33}/(a_{22}a_{33} - a_{23}^2) & a_{23}/(a_{22}a_{33} - a_{23}^2) \\ 0 & a_{23}/(a_{22}a_{33} - a_{23}^2) & a_{22}/(a_{22}a_{33} - a_{23}^2) \end{bmatrix} \quad (27)$$

So, finally, the optimization problem (15) is reduced to

$$O_t = \frac{d_1}{a_{11}} + \frac{d_2 a_{22} + d_3 a_{33} + 2 d_4 a_{23}}{a_{22}a_{33} - a_{23}^2} \to \min_{q_{2i}, q_{3i}, \alpha_i} \quad (28)$$

where the coefficients $d_1, d_2, d_3, d_4$ are defined by the test configuration $\mathbf{q}^0 = (q_1^0, q_2^0, q_3^0)$ and the external loading $\mathbf{F}^0 = [F_x^0 \ F_y^0 \ F_z^0]^T$. These coefficients can be computed via the columns of the matrix $\mathbf{A}^0$ as

$$d_1 = (\mathbf{A}_1^0)^T \mathbf{A}_1^0; \quad d_2 = (\mathbf{A}_3^0)^T \mathbf{A}_3^0; \quad d_3 = (\mathbf{A}_2^0)^T \mathbf{A}_2^0; \quad d_4 = (\mathbf{A}_2^0)^T \mathbf{A}_3^0; \quad (29)$$

Therefore,

$$\begin{aligned}
d_1 &= (l_C^0)^4 \left( F_x^0 \sin(q_1^0) - F_y^0 \cos(q_1^0) \right)^2 \\
d_2 &= (l_3)^2 \left( F_x^0 l_3 \sin(q_2^0 + q_3^0) \cos(q_1^0) + F_y^0 l_3 \sin(q_2^0 + q_3^0) \sin(q_1^0) - F_z^0 l_3 \cos(q_2^0 + q_3^0) \right)^2 \\
d_3 &= \left( F_x^0 l_S^0 \cos(q_1^0) + F_y^0 l_S^0 \sin(q_1^0) - F_z^0 l_C^0 \right)^2 \cdot \left( l_2^2 + l_3^2 + 2 l_2 l_3 \cos(q_3^0) \right) \\
d_4 &= l_3 \left( l_3 + l_2 \cos(q_3^0) \right) \left( F_x^0 l_S^0 \cos(q_1^0) + F_y^0 l_S^0 \sin(q_1^0) - F_z^0 l_C^0 \right) \cdot \\
&\quad \cdot \left( F_x^0 l_3 \sin(q_2^0 + q_3^0) \cos(q_1^0) + F_y^0 l_3 \sin(q_2^0 + q_3^0) \sin(q_1^0) - F_z^0 l_3 \cos(q_2^0 + q_3^0) \right)
\end{aligned} \quad (30)$$

It is evident that the optimization problem (28) does not have a trivial analytical solution. Nevertheless, since test configuration $\mathbf{q}^0 = (q_1^0, q_2^0, q_3^0)$ and external loading $\mathbf{F}^0 = (F_x^0, F_y^0, F_z^0)^T$ are defined, it is possible to obtain a numerical solution. In order to illustrate the efficiency of the proposed approach, numerical simulations have been carried out for one, two, three and four measurements of the end-effector deflections under the test loading. Modeling results for $l_1 = 0.75 \ m$, $l_2 = 1.25 \ m$, $l_3 = 1.10 \ m$, $\mathbf{q}^0 = (0°, 60°, -45°)$, $\mathbf{F}^0 = F_0 [0, 0.29, -0.96]^T$ are summarized in Table 1. They include the quality of the experimental configurations (performance measure), calibration configurations and identification accuracy for the joint stiffnesses. For comparison purposes, the results have been obtained using three different plans of calibration experiments: (i) calibration in the test configuration, (ii) calibration in the optimal configuration that has been obtained for the case of one experiment and (iii) calibration in the optimal configurations that has been obtained using (28).

These results show that the proposed test-pose optimization criterion improves the efficiency of the compliance errors compensation by a factor of two comparing to calibration in the test configuration. Besides, it improves the identification accuracy of the joint compliances, so obtained results also insure better end-point positioning accuracy in other configurations.

It should be stressed that carrying out several experiments in the optimal configuration obtained with one experiment gives identification accuracy close to the optimal plan. So, in practice, when it is complicated to change the robot configuration for each experiment it is possible to carry out experiments in one configuration obtained for identification of elastostatic parameters from one experiment. This approach reduces the identification accuracy by 20%, however an additional experiment may compensate this loss of the accuracy.

Table 1   Calibration of elastostatic parameters using different plans of experiments

| Case studies | Performance measure | Calibration configuration | | | Identification accuracy, [rad/N] | | |
|---|---|---|---|---|---|---|---|
| | | $q_2$ | $q_3$ | $\alpha$ | $\delta k_1$ | $\delta k_2$ | $\delta k_3$ |
| Test Conf. | 3.00 $\sigma^2$ | 60° | 45° | -73.3° | 1.22 $\sigma$ | 0.70 $\sigma$ | 2.19 $\sigma$ |
| Opt.$_1$ Conf. | 1.92 $\sigma^2$ | 43.2° | -57.3° | 22.9° | 0.66 $\sigma$ | 0.52 $\sigma$ | 1.81 $\sigma$ |
| 2×Test Conf. | 1.50 $\sigma^2$ | 60° | 45° | -73.3° | 0.86 $\sigma$ | 0.49 $\sigma$ | 1.55 $\sigma$ |
| 2×Opt.$_1$ Conf. | 0.96 $\sigma^2$ | 43.2° | -57.3° | 22.9° | 0.47 $\sigma$ | 0.37 $\sigma$ | 1.28 $\sigma$ |
| Opt.$_2$ Conf. | 0.80 $\sigma^2$ | 5.5° | -6.8° | 26.3° | 0.41 $\sigma$ | 0.30 $\sigma$ | 0.96 $\sigma$ |
| | | 93.1° | -101.2° | 3.3° | | | |
| 3×Test Conf. | 1.00 $\sigma^2$ | 60° | 45° | -73.3° | 0.71 $\sigma$ | 0.40 $\sigma$ | 1.27 $\sigma$ |
| 3×Opt.$_1$ Conf. | 0.64 $\sigma^2$ | 43.2° | -57.3° | 22.9° | 0.38 $\sigma$ | 0.30 $\sigma$ | 1.05 $\sigma$ |
| Opt.$_3$ Conf. | 0.51 $\sigma^2$ | 173.3° | 19.3° | 0.5° | 0.32 $\sigma$ | 0.23 $\sigma$ | 0.83 $\sigma$ |
| | | -7.1° | 14.7° | -24.9° | | | |
| | | -49.3° | -125.0° | 2.1° | | | |
| 4×Test Conf. | 0.75 $\sigma^2$ | 60° | 45° | -73.3° | 0.61 $\sigma$ | 0.35 $\sigma$ | 1.10 $\sigma$ |
| 4×Opt.$_1$ Conf. | 0.48 $\sigma^2$ | 43.2° | -57.3° | 22.9° | 0.33 $\sigma$ | 0.26 $\sigma$ | 0.91 $\sigma$ |
| Opt.$_4$ Conf. | 0.39 $\sigma^2$ | 28.3° | -39.1 | 9.7° | 0.25 $\sigma$ | 0.21 $\sigma$ | 0.78 $\sigma$ |
| | | 4.6° | -12.6° | 22.4° | | | |
| | | -3.4° | -4.8° | -37.4° | | | |
| | | 146.8° | -150.6° | -5.2° | | | |

Test Conf. - Calibration in the test configuration ( $\mathbf{q}^0 = (0°, 60°, -45°)$ , $\mathbf{F}^0 = F_0[0, 0.29, -0.96]^T$ )
Opt.$_1$ Conf. - Calibration in the optimal configuration obtained with one experiment
Opt.$_2$ Conf. - Calibration in the optimal configuration obtained with two experiments
Opt.$_3$ Conf. - Calibration in the optimal configuration obtained with three experiments
Opt.$_4$ Conf. - Calibration in the optimal configuration obtained with four experiments
Other parameters $q_1 = 0$ , $l_1 = 0.75\ m$ , $l_2 = 1.25\ m$ , $l_3 = 1.10\ m$ , $F_0$ defines by a user

**Summary**

The paper presents a new approach for design of elastostatic calibration experiments that allows essentially reducing the identification errors due to proper selection of the manipulator postures employed in the measurements. In contrast to other works, the quality of the measurement configurations is estimated using a new test-pose based optimization criterion that allows to combine multiple objectives with different units in a single performance measure. This approach increases the efficiency of the compliance error compensation and ensures the best position accuracy for the considered test configuration under the task loading. The proposed criterion can be treated as the weighted trace of the covariance matrix, where the weighting coefficients are derived using the test pose. Validity of the obtained results and their practical significance were confirmed by means of a simulation study that deals with 3-link anthropomorphic robot.

**Acknowledgments**

The authors would like to acknowledge the financial support of the ANR, France (Project ANR-2010-SEGI-003-02-COROUSSO) and the Region "Pays de la Loire", France.